\documentclass[10pt,twocolumn,letterpaper]{article}

\usepackage[pagenumbers]{cvpr} 

\usepackage[utf8]{inputenc} 
\usepackage[T1]{fontenc}    
\usepackage{url}            
\usepackage{booktabs}       
\usepackage{amsfonts}       
\usepackage{nicefrac}       
\usepackage{microtype}      
\usepackage{xcolor}         
\usepackage{graphicx}
\usepackage{algorithm}
\usepackage{algorithmic}
\usepackage{amsmath,amssymb,mathtools,amsthm}
\usepackage{pifont}
\usepackage{cuted}
\usepackage[most]{tcolorbox}
\newcommand{\cmark}{\text{\ding{51}}}%
\newcommand{\xmark}{\text{\ding{55}}}%

\newcommand{\gxmark}{\textcolor{gray}{\xmark}}
\newcommand{\vc}{\ensuremath{\boldsymbol{c}}}
\newcommand{\xw}{\ensuremath{\mathbf{x}^w}}
\newcommand{\xl}{\ensuremath{\mathbf{x}^l}}
\newcommand{\x}{\ensuremath{\mathbf{x}}}
\newcommand{\pref}{p_{\text{ref}}}

\definecolor{cvprblue}{rgb}{0.21,0.49,0.74}
\usepackage[pagebackref,breaklinks,colorlinks,allcolors=cvprblue]{hyperref}

\title{DesignDiffusion: High-Quality Text-to-Design Image Generation \\with Diffusion Models}

\author{Zhendong Wang$^{1}$, Jianmin Bao$^{2,*,\dagger}$, Shuyang Gu$^{2}$, Dong Chen$^{2}$, Wengang Zhou$^{1,3,*}$, Houqiang Li$^{1,3}$ \\
\small$^{*}$corresponding author \qquad $^{\dagger}$project lead \\
$^{1}$ CAS Key Laboratory of GIPAS, EEIS Department, University of Science and Technology of China \\
$^{2}$ Microsoft Research Asia \\
$^{3}$ Institute of Artificial Intelligence, Hefei Comprehensive National Science Center
}

\begin{document}
\maketitle

\begin{strip}
\centering
\vspace{-1.7cm}
\includegraphics[width=1\linewidth]{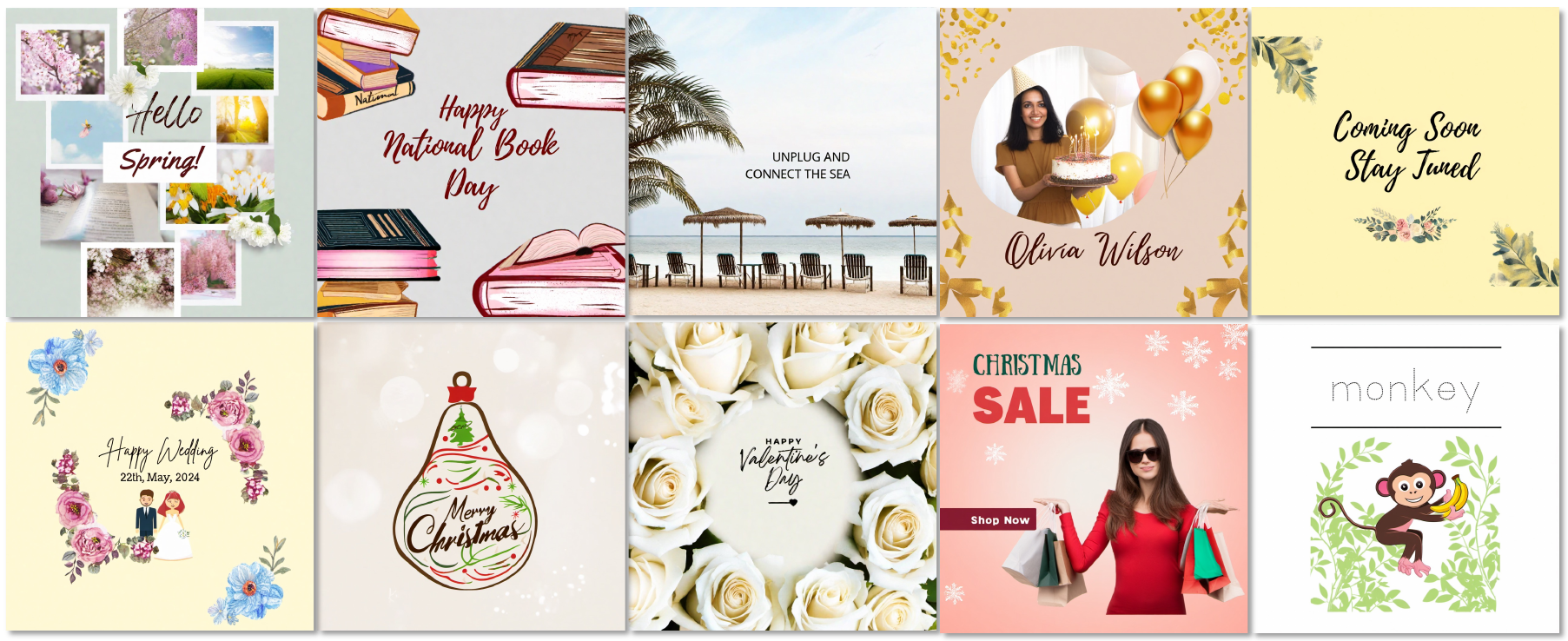}
\vspace{-2em}
\captionof{figure}{Images generated by our DesignDiffusion model, which only requires a simple text prompt as input and can generate diverse, high-quality design images with accurate textual and vivid visual content. \label{fig:teaser}}
\end{strip}

\begin{abstract}
In this paper, we present DesignDiffusion, a simple yet effective framework for the novel task of synthesizing design images from textual descriptions. A primary challenge lies in generating accurate and style-consistent textual and visual content. Existing works in a related task of visual text generation often focus on generating text within given specific regions, which limits the creativity of generation models, resulting in style or color inconsistencies between textual and visual elements if applied to design image generation. To address this issue, we propose an end-to-end, one-stage diffusion-based framework that avoids intricate components like position and layout modeling. Specifically, the proposed framework directly synthesizes textual and visual design elements from user prompts. It utilizes a distinctive character embedding derived from the visual text to enhance the input prompt, along with a character localization loss for enhanced supervision during text generation. Furthermore, we employ a self-play Direct Preference Optimization fine-tuning strategy to improve the quality and accuracy of the synthesized visual text. Extensive experiments demonstrate that DesignDiffusion achieves state-of-the-art performance in design image generation.
\end{abstract}

\section{Introduction}
\label{sec:intro}

Design images play a pivotal role in various applications, such as graphic design, advertising, and scientific visualization, where the integration of visual and textual elements must be flawless. 

Recently, many fascinating and wonderful applications, such as art image creation or film production, have shown remarkable progress due to the rapid advance of text-to-image generation techniques~\cite{ramesh2021zero,LDM,saharia2022photorealistic,ruiz2022dreambooth,zhang2023adding}.
Despite these advancements, the precise generation of design images poses a distinct challenge, largely due to the limited capabilities of existing image generation models in complex layout planning and visual text generation.

To address the critical issue of visual text generation, recent studies~\cite{ tuo2023anytext, liu2024glyph} have made significant contributions. A common approach involves a two-step process: first generating an image from textual input and then identifying a suitable location within the image to insert text. However, this method presents several challenges. Firstly, a generated image may not have a suitable region for adding text, resulting in an overlay of text on visual elements. Secondly, even with a plausible location, the rendered text may lack consistency with the visual elements presented in the image, compromising overall coherence and professionalism.

An alternative research direction in visual text rendering, as suggested by ~\cite{chen2023textdiffuser, textdiffuser-2}, focuses on learning the layout of text regions from textual prompts in advance. Then, this learned layout serves as a condition to guide text-to-image models in generating text within designated regions of the image. However, the dependence on predefined text regions might constrain the creative breadth of the generated images. Moreover, these predetermined text regions may not always adhere to natural design principles, potentially leading to visually unattractive or conceptually incoherent outputs.

To address the aforementioned issues, we propose DesignDiffusion, an end-to-end diffusion-based framework specifically developed for the novel task of generating design images from text. The holistic learning approach presents multiple benefits: (1) it eliminates the need to predetermine text regions in the generative images, thereby preserving the creative freedom of the generative models; (2) it enables the smooth incorporation of generated text into images, yielding cohesive designs that accurately reflect input prompts.

In our framework, we employ three essential techniques when fine-tuning text-to-image diffusion models:
a) Recognizing that a conventional CLIP text encoder might overlook the specific textual content designated for rendering in the generated image, we propose a character-level decomposition of words. This approach provides the CLIP text encoder hints to render each character in the image precisely.
b) To encourage the model to pay more attention to newly introduced character embeddings, we propose a character localization loss. This loss function compels each character embedding to concentrate on its respective area within the image, ensuring an accurate visual representation of each character.
c) We introduce a new training strategy called Self-Play Direct Preference Optimization~(SP-DPO) for fine-tuning the model to enhance its capability of generating accurate and high-fidelity text. SP-DPO is based on the hypothesis that ground truth data are typically preferred by humans over model-generated samples. Based on this, we fine-tune the model to align with these human preferences.

Extensive experiments demonstrate the superior capabilities of DesignDiffusion for design image generation. It generates creative design images with prompts, as shown in Figure~\ref{fig:teaser}. Quantitative and qualitative analysis shows that our approach significantly surpasses current state-of-the-art text-to-image and visual text rendering models in terms of both image quality and text accuracy.

In summary, our contributions are presented as follows:
\begin{itemize}
    \item We propose DesignDiffusion, an end-to-end, diffusion-based framework for text-to-design image generation, which enables the simultaneous generation of image and visual text elements, eliminating the necessity for predefined text regions or traditional two-stage separated text and image creation process.
    \item We explore prompt enhancement for CLIP text encoding along with a character localization loss, to improve the accuracy of text placement and the fidelity of text rendering.
    \item By adopting an SP-DPO mechanism, the generative model is further fine-tuned, achieving remarkable improvements in visual text accuracy and overall image quality.
\end{itemize}

\begin{figure*}[t] 
    \centering 
    \includegraphics[width=1.0\linewidth]{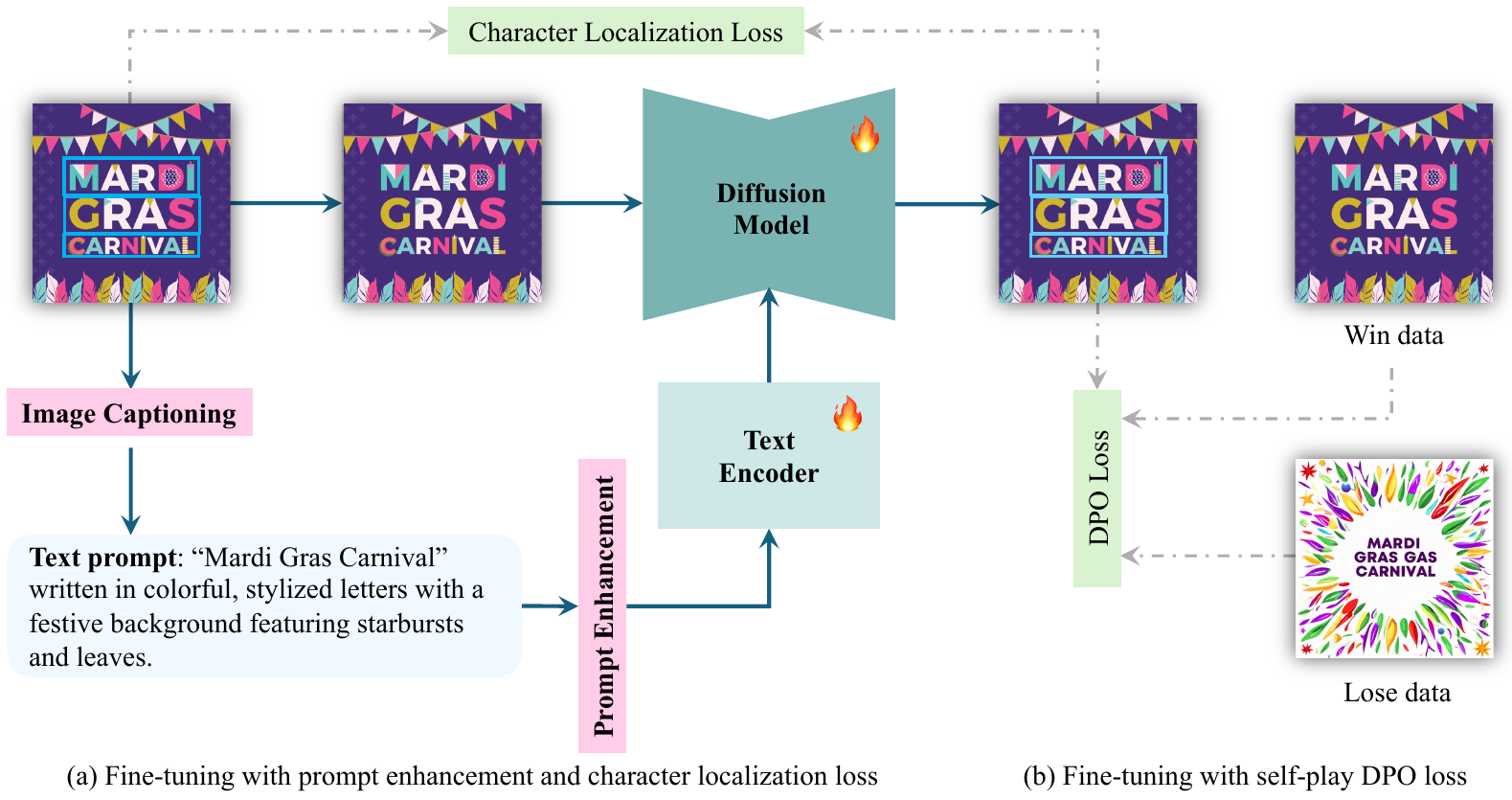}
    \vspace{-1.5em}
    \caption{Overview of the DesignDiffusion framework. DesignDiffusion is based on enhanced text prompts, with trainable CLIP text encoder and UNet, and does not require additional complex conditions~(glyphs, positions). Character localization loss is added as extra supervision at cross-attention maps to force the UNet to attend more to the visual character regions. To further improve the quality of visual text generation, we incorporate a self-play DPO strategy into the fine-tuning process. Diffusion denoise loss is omitted here.}
    \label{fig:framework}
    \vspace{-1em}
\end{figure*}

\section{Related Work}
\label{sec:related_work}
\noindent\textbf{Visual Text Rendering.} Despite the remarkable success of existing diffusion models~\cite{ramesh2021zero, LDM, saharia2022photorealistic} in text-to-image generation, visual text rendering remains a challenging task. Recently, researchers have recognized this issue and proposed several approaches to address it, such as GlyphDraw~\cite{ma2023glyphdraw}, DiffUTE~\cite{chen2024diffute}, GlyphControl~\cite{yang2023glyphcontrol}, AnyText~\cite{tuo2023anytext}, TextDiffuser series~\cite{chen2023textdiffuser, textdiffuser-2}, and \cite{liu2022character}. While these efforts have improved text rendering accuracy, it is worth noting that these models obtain limited image quality due to the need of pre-giving text rendering areas during the generation process or inpainting text into a well-generated image. Separating text and visual content generation away seriously limits the overall quality of generated images with visual text rendered, which is incompatible with the goal of design image generation.

\noindent\textbf{Compositional Image Generation.} To improve the ability of text-to-image models to create complex compositions, researchers have integrated new components into diffusion models, enabling the generation of images with multiple features. Examples include T2I-Adapter~\cite{mou2023t2i} and ControlNet~\cite{zhang2023adding}, which incorporate semantic details like layout and pose to influence the structure of images~\cite{zhang2023controllable}. In a similar vein, GLIGEN~\cite{li2023gligen} utilizes adapters with diffusion models to achieve image generation that is conditioned on specific spatial arrangements. However, these methods cannot generate images directly from prompts and require substantial human effort to define the image layout. On the other hand, some studies have investigated methods that do not require additional training to prompt diffusion models to create images based on spatial conditions during the inference phase. Recent advancements~\cite{lian2023llm, pan2023kosmos, yang2024mastering} have seen large language models aiding diffusion models in generating images with intricate or predefined layouts. Yet, these approaches often fall short in generating design-centric images that seamlessly integrate both textual and visual components in a well-organized manner.

\noindent\textbf{Aligning Diffusion Models.} Replicating the success of reinforcement learning in large language models~\cite{ouyang2022training, rafailov2023direct} to text-to-image diffusion models remains a significant challenge. Recent works have proposed various methods to address this problem. For example, EMU~\cite{dai2023emu} identifies that a few high-quality samples are crucial when fine-tuning the model to generate visually appealing images. DPOK~\cite{fan2024reinforcement} and DDPO~\cite{ddpo} use RL-based approaches to maximize rewards on a limited vocabulary set, performing well when the number of train/test prompts is small. Another work DiffusionDPO~\cite{wallace2023diffusion} applies DPO based on a human-annotated preference dataset. Our DesignDiffusion differs by applying a self-play DPO strategy, which does not require human annotation and lacks of exploration in design image generation.

\section{DesignDiffusion}
\label{sec:method}
The generative community has witnessed the strong generative capability of diffusion models~\cite{podell2023sdxl,LDM,DDPM,ADM,zhang2023controllable}. However, generating design images remains a great challenge, particularly in the context of exquisite layout and text rendering. Previous approaches~\cite{textdiffuser-2,chen2023textdiffuser} in a related task~(visual text rendering) attempt to inpaint text after generating an image, or to plan text regions using large language models such as Vicuna~\cite{zheng2023judging} and GPT~\cite{achiam2023gpt} before generating an image. We argue that they do not consider the integration of text and image content. Note that, the text and image contexts in design images are often intertwined and inseparable, such that they are incompatible with design image generation.

In contrast, we introduce a straightforward, one-stage approach for text-to-design image generation, relying solely on the user prompt. It is based on an aesthetic model, \ie, Stable Diffusion XL (SDXL) model~\cite{podell2023sdxl},  incorporating three core components that enable the successful generation of text-rich design images. An overview of the DesignDiffusion framework is provided in Fig.~\ref{fig:framework}. The DesignDiffusion training involves a two-stage fine-tuning strategy. In the first stage, we introduce prompt enhancement and character localization loss into the framework to generate high-quality images with accurate text. In the second stage, we propose SP-DPO to further improve the text rendering accuracy.
These components are detailed in the following subsections.

\begin{algorithm*}[t]
    \caption{Self-play DPO fine-tuning, given a base diffusion model $\mathbf{\epsilon}_{\text{ref}}$, a trainable copy $\mathbf{\epsilon}_{\theta}$, a prompt-image paired design-image dataset $\mathbf{D}=\{\mathbf{x}_i,\mathbf{y}_i\}^{N}_{i=1}$, in which $\mathbf{x}_i$ and $\mathbf{y}_i$ represent image and prompt, respectively.}
    \label{alg:SPIN_DPO}
    \begin{algorithmic}
        \FORALL{$i$ from 1 to $N$}
            \STATE Generate $K$ candidate samples $\{g_{ik}\}_{k=1}^{K}$ using $\mathbf{\epsilon}_{\text{ref}}$ with input $\mathbf{y}_i$
            \STATE Losing data: $\mathbf{x}_i^l = g_{ik}$, in which $g_{ik}$ gets the worst text rendering accuracy
            \STATE Winning data: $\mathbf{x}_i^w = \mathbf{x}_i$
            
        \STATE Update $\mathbf{\epsilon}_{\theta}$ by: $\min\limits_{\mathbf{\epsilon}_{\theta}} \sum\limits_{i = 0}^{n}(\text{log}\sigma(-\beta T\omega(\lambda_t)) (\| \epsilon_i^w -\epsilon_\theta(\mathbf{x}_{i,t}^w,\mathbf{y}_i,t_i)\|^2_2 -  $
    \STATE~~~~~~ $ \|\epsilon_i^w - \epsilon_\text{ref}(\mathbf{x}_{i,t}^w,\mathbf{y}_i,t_i)\|^2_2 - ( \| \epsilon_i^l -\epsilon_\theta(\mathbf{x}_{i,t}^l,\mathbf{y}_i,t_i)\|^2_2 - \|\epsilon_i^l - \epsilon_\text{ref}(\mathbf{x}_{i,t}^l,\mathbf{y}_i,t_i)\|^2_2))) $
        \ENDFOR
        \RETURN $\mathbf{\epsilon}_{\theta}$
    \end{algorithmic}
\end{algorithm*}

\subsection{Prompt Enhancement for CLIP Text Encoding}
\label{subsec:char_emb}

Currently, available text-to-image diffusion models struggle to generate text-rich design images based solely on an input text prompt. This challenge arises from two main issues: 1) Visually rendered text must be letter-by-letter correct, however, during tokenization, most words are tokenized as a single token, ignoring each letter within the word thus complicating the learning of individual letters. 2) The text in images is often diverse in style, size, and layout, increasing the complexity of visual text modeling. To address the first issue, we introduce a novel approach by embedding each letter from prompts directly into CLIP text encoders within latent diffusion models~\cite{podell2023sdxl, LDM}. Specifically, the input prompt of the CLIP text encoders is enhanced by appending an extra text description to the original prompt. The template for the text description is ``Rendered characters: <|startofchar|> $\cdots$ <|endofchar|>". For example, as illustrated in Fig.~\ref{fig:framework}, if the desired rendered text is ``Mardi Gras Carnival'', the appended text description would be ``Rendered characters:<|startofchar|> <|M|> <|a|> <|r|> <|d|> <|i|><|endofchar|><|startofchar|> <|G|> <|r|> <|a|> <|s|> <|endofchar|> <|startofchar|> <|C|> <|a|> <|r|> <|n|> <|i|> <|v|> <|a|> <|l|><|endofchar|>''. A character surrounded by `|' denotes a newly introduced trainable token.

We introduce a total of 97 new tokens, encompassing 26 uppercase letters, 26 lowercase letters, 10 numbers, 32 punctuation marks, a space, a start flag, and an end flag. The enhanced prompt enables the CLIP text encoder to be trained specifically to learn hints for rendering text. By augmenting the prompt with detailed rendered text information, the UNet learns to generate images that include the specified visual text content seamlessly integrated into the image, without relying on predefined text glyphs or positions. This approach addresses the challenge of integrating text and image content in a tightly coupled manner, distinguishing it from previous approaches that treated text and image generation as separate stages. This integrated technique ensures that design images with visual text are generated in a more natural and contextually appropriate way.

\subsection{Localizing Cross-Attention Maps with Character Segmentation Masks}
\label{subsec:char_loss}

To address the second issue, \ie, complexity (\eg, style, size, color, and orientation) of text in images, we explore applying a text localization loss during training. This approach helps the model capture text effectively. Previous works~\cite{hertz2022prompt, xiao2023fastcomposer} have shown that cross-attention maps in the UNet component of latent diffusion models~\cite{LDM, podell2023sdxl} control which pixels attend to which tokens of the prompt text during diffusion generation. In our framework, the newly introduced trainable character embeddings in CLIP text encoders are not involved in the original UNet pre-training. Consequently, the cross-attention maps between rendered character tokens and latent pixel features may be unreasonable. To encourage UNet to attend to the newly introduced character embeddings, we propose a character localization loss to associate each character token with its corresponding visual character region. Let $\mathbf{A}^{(h \times w \times c)}$ denote the cross-attention maps, where $h$ and $w$ are the spatial dimensions of the latent in Stable Diffusion, respectively. $\mathbf{A}^{i} = \mathbf{A}[:,:,i]$ represents the relationship between the image features and the $i$-th prompt token. $\mathbf{M}^{i}$ is the region mask of the $i$-th prompt token, with 1 denoting the existence of elements. Encouraging each character token to attend to the corresponding character region, the character localization loss is formulated as follows:
\vspace{-0.5em}
\begin{equation}
\mathcal{L}_{char} = -\frac{1}{c}\sum_{i=0}^{c}(\mathbf{A}^{i}*\mathbf{M}^{i} -\mathbf{A}^{i}*(1-\mathbf{M}^{i})).
\label{eq:char_loss}
\end{equation}

This character localization loss is used as a regularization term that forces the generation of visual characters to attend to the corresponding character embedding. We use the cross-attention maps in the five middle stages of UNet to calculate $\mathcal{L}_{char}$ in our experiments.

\subsection{Self-Play DPO Fine-Tuning for More Accurate Text Rendering}
\label{subsec:sp-dpo}
Recent diffusion DPO~\cite{rafailov2024direct} is an alternative to RLHF to steer the model to the preferred distribution~(winning data) and push it away from the unpreferred distribution~(losing data). 
To further enhance the model's ability to handle design images, we apply self-play DPO fine-tuning with our training GT images as winning data and synthetic data as losing data. Specifically, we generate losing data from the preliminary fine-tuned model in the first stage. To avoid high-quality generated images being mistakenly regarded as losing samples, we generate $K$ samples for each prompt. Subsequently, we use a large multi-modality model~\cite{liu2023llava, liu2023improvedllava} to extract the rendered text in images. The images with the poorest rendering quality are retained as the final losing data. The DPO reward objective is defined as follows: 
\vspace{-0.5em}
\begin{equation}
\begin{split}
    L_{\text{DPO}}(\theta)\!=\!-
    \mathbb{E}_{\vc,\xw_0,\xl_0}[
    \log\sigma(\beta \log \frac{p_{\theta}(\xw_0|\vc)}{\pref(\xw_0|\vc)}\\-
    \beta \log \frac{p_{\theta}(\xl_0|\vc)}{\pref(\xl_0|\vc)})],
    \label{eq:dpo_reward}
\end{split}
\end{equation}
where $\vc$ represents the condition, $\xw_0$ and $\xl_0$ denote the winning and losing data, respectively. $\beta$ is a hyperparameter. $\pref$ is a reference distribution, and $p_{\theta}$ is the optimized distribution with network parameters $\theta$. 

Applying this loss to diffusion, $p_{\theta}(\x_0|c)$ is not tractable, as it needs to marginalize out all possible diffusion paths ($\x_1,\dots,\x_T$ ) which leads to $\x_0$. So, we utilize the evidence lower bound (ELBO) to overcome this challenge. Specifically, we need to redefine the reward function on the whole chain ($\x_0,\x_1,\dots,\x_T$). The Eq.~\ref{eq:dpo_reward} becomes to Eq.~\ref{eq:dpo_elbo}:
\vspace{-0.5em}
\begin{equation}
\begin{split}
L_{\text{Diffusion-DPO}}(\theta)=-E_{x_0^w,x_0^l}\log\sigma(\beta E_{x_{0:T}^w,x_{0:T}^l}\\
[\log \frac{p_{\theta}(x_{0:T}^w)}{p_{\text{ref}}(x_{0:T}^w)}-\log \frac{p_{\theta}(x_{0:T}^l)}{p_{\text{ref}}(x_{0:T}^l)}]),\label{eq:dpo_elbo}
\end{split}
\end{equation}
$c$ is omitted for simplification. Since $x_{1:T}\sim p_{\theta}(x_{1:T}|x_0)$, T is usually large, and $p_{\theta}(x_{1:T}|x_0)=p_T \prod \limits_{t=1}^T p_{\theta}(x_{t-1}|x_t)$. Considering Jensen’s inequality and convexity, we can further get the upper bound of Eq.~\ref{eq:dpo_elbo}:
\vspace{-0.5em}
\begin{equation}
\begin{split}
    L_{\text{Diffusion-DPO}}(\theta)<-E_{x_0^w,x_0^l,t,x_{t-1}^w\sim p_{\theta}(x_{t-1}^w|x_0^w),x_{t-1}^l\sim p_{\theta}(x_{t-1}^l|x_0^l)} \\
    \log\sigma(\beta T\log \frac{p_{\theta}(x_{t-1}^w|x_t^w)}{p_{\text{ref}}(x_{t-1}^w|x_t^w)}-\beta T \log \frac{p_{\theta}(x_{t-1}^l|x_t^l)}{p_{\text{ref}}(x_{t-1}^l|x_t^l)}).\label{eq:dpo_jensen}
\end{split}
\end{equation}
Approximate the reverse process $p_{\theta}(x_{1:T}|x_0)$ with the forward process $q(x_{1:T}|x_0)$, Eq.~\ref{eq:dpo_jensen} becomes:
\vspace{-0.5em}
\begin{equation}
\begin{split}
    L(\theta)=-E_{x_0^w,x_0^l,t,x_t^w\sim q_(x_t^w|x_0^w),x_t^l\sim q(x_t^l|x_0^l)}
    \log\sigma (\beta T(\\D_{\text{KL}}q(x_{t-1}^w|x_{0,t}^w)||p_{\theta}(x_{t-1}^w|x_t^w)\\-D_{\text{KL}}q(x_{t-1}^w|x_{0,t}^w)||p_{\text{ref}}(x_{t-1}^w|x_t^w)\\-D_{\text{KL}}q(x_{t-1}^l|x_{0,t}^l)||p_{\theta}(x_{t-1}^l|x_t^l)\\+D_{\text{KL}}q(x_{t-1}^l|x_{0,t}^l)||p_{\text{ref}}(x_{t-1}^l|x_t^l))).
\end{split}
\end{equation}
\vspace{-0.5em}
Given the diffusion reverse process, the final objective is:
\vspace{-0.5em}
\begin{equation}
\begin{split}
    L(\theta)=-E_{x_0^w,x_0^l,t,x_t^w\sim q_(x_t^w|x_0^w),x_t^l\sim q(x_t^l|x_0^l)}
    \log\sigma(-\beta T\omega(\lambda_t) \\(||\epsilon^w-\epsilon_{\theta}(x_t^w,t)||^2_2- ||\epsilon^w-\epsilon_{\text{ref}}(x_t^w,t)||^2_2\\-||\epsilon^l-\epsilon_{\theta}(x_t^l,t)||^2_2+||\epsilon^l-\epsilon_{\text{ref}}(x_t^l,t)||^2_2)),
\end{split}
\end{equation}
in which $\omega(\lambda_t)$ is a timestep-wise weighting function, which is constant in common practice. Finally, the overall self-play DPO fine-tuning algorithm is described in Algorithm~\ref{alg:SPIN_DPO}.

\begin{table*}[t]
\caption{Quantitative comparisons with state-of-the-art text-to-image generation and visual text generation methods demonstrate that DesignDiffusion achieves superior performance in terms of FID, text precision, recall, F-score, and accuracy.}
\vspace{-1em}
\centering
\resizebox{1.0\linewidth}{!}{

\setlength\tabcolsep{15pt}
\begin{tabular}{l|ccccc}
\hline
Method & FID$\downarrow$  & Text Precision$\uparrow$ & Text Recall$\uparrow$ & Text F-score$\uparrow$ & Text Accuracy$\uparrow$\\ \hline
SDXL~\cite{podell2023sdxl} & 45.10  & 0.517 & 0.461 & 0.488 & 0.241 \\
SD3~\cite{sd3} & 45.90  & 0.754 & 0.675 & 0.712 & 0.415\\
FLUX~\cite{flux2023} & 35.28 & 0.764 & 0.698 & 0.729 & 0.442 \\
DeepFloyd-IF~\cite{deepfloyd} & 52.05 & 0.607 & 0.534 & 0.568 & 0.290\\
GlyphControl~\cite{yang2024glyphcontrol} & 60.07  & 0.462 & 0.352 & 0.400 & 0.218 \\
Textdiffuser~\cite{chen2023textdiffuser} & 55.83 & 0.806 & 0.706 & 0.752 & 0.478 \\
Textdiffuser-2~\cite{textdiffuser-2} & 63.92 & 0.866 & 0.792 & 0.827 & 0.583 \\
AnyText~\cite{tuo2023anytext} & 64.69 & 0.440 & 0.378 & 0.407 & 0.230\\
\textbf{DesignDiffusion} & \textbf{19.87} & \textbf{0.888} & \textbf{0.837} & \textbf{0.862} & \textbf{0.631} \\ \hline
\end{tabular}
}
\vspace{-1em}
\label{table:quan_result}
\end{table*}

\begin{table*}[t]
\caption{Results of a user study comparing our method with previous approaches based on four aspects of text and image quality.}
\vspace{-1em}
\label{table:user_study}
\centering
\resizebox{1.0\textwidth}{!}{
\setlength\tabcolsep{12pt}
\begin{tabular}{l|cccc}
\hline
Method & Image Aesthetic$\uparrow$ & Text Quality $\uparrow$ & Layout Aesthetics$\uparrow$ &Text-Image Matching$\uparrow$ \\ \hline
SDXL~\cite{podell2023sdxl}& \textbf{0.43} & 0.11& 0.18 & 0.28\\
Textdiffuser-2~\cite{textdiffuser-2} & 0.08& 0.34& 0.22& 0.14\\
AnyText~\cite{tuo2023anytext} & 0.09&0.03 & 0.07 & 0.06\\
DesignDiffusion & 0.40 & \textbf{0.51}& \textbf{0.54} & \textbf{0.52} \\ \hline
\end{tabular}
}
\vspace{-1em}
\end{table*}

\section{Experiments}
\label{sec:exp}
\subsection{Settings}
\noindent\textbf{Dataset.} Due to the lack of high-quality design image dataset, we collect a dataset of 1 million high-quality design images, including logos, posters, flyers, covers, etc. The collected dataset contains region masks that denote image layer information and text annotations including text content and font size in images. For image captioning, we use a large vision-language model LLaVA 1.6-34B~\cite{liu2023llava,liu2023improvedllava,liu2024llavanext} with improved reasoning, OCR, and world knowledge. 
Given the output from LLaVA, we improve it by involving the annotated text information.
Computing auxiliary localization loss for characters requires a character-wise region annotation. Applying a pre-trained character segmentation model~\cite{chen2023textdiffuser}, we obtain character segmentation images. Considering text annotation and connected graph analysis, final character-wise region masks are obtained. In summary, our training dataset contains 1 million high-quality design images with caption, text, and character region annotations. Our testing dataset comprises 5$k$ samples, keeping not seen during training.
For more details about the dataset, please refer to Appendix.

\noindent\textbf{Training details.} Our DesignDiffusion is fine-tuned on StableDiffusion XL base model~\cite{podell2023sdxl}.
It is trained using AdamW~\cite{loshchilov2017decoupled} with a learning rate of 1e-5. Xformers~\cite{xFormers2022} and DeepSpeed~\cite{parmar2023zero} are applied during training to save memory further. Our model is trained on 4$\times$NVIDIA A100 GPUs with a batch size of 256 in all our experiments. The coefficient $\lambda$ for $\mathcal{L}_{char}$ is 0.05 by default. After that, for DPO fine-tuning, we use 1.5$k$ win-lose pairs as training data. $\beta$ of DPO and $T$ in Algorithm~\ref{alg:SPIN_DPO} are set as 5 and 1,000, respectively. The learning rate is 1e-8 and the number of candidate samples $K=10$ in DPO experiments. During the training of both stages, classifier-free guidance~\cite{ho2022classifier} is applied with a probability of 10\%. When inference, the classifier-free guidance scale is set to 7.5 by default.

\noindent\textbf{Evaluation metrics.} Text-to-design image generation evaluation focuses on both image quality and text quality evaluation. (1) We apply \textbf{FID}~\cite{heusel2017gans} to evaluate the generated image distribution and the real design image distribution. 
(2) For evaluating the text quality in generated design images, we report \textbf{OCR} precision, recall, F-Score, and accuracy. Please refer to Appendix for a more detailed analysis of OCR tools.
(3) Besides the evaluation from available tools or metrics, \textbf{human evaluation} is used in our experiments to evaluate the generated image and text quality comprehensively.

\begin{figure*}
  \centering
  \includegraphics[width=\linewidth]{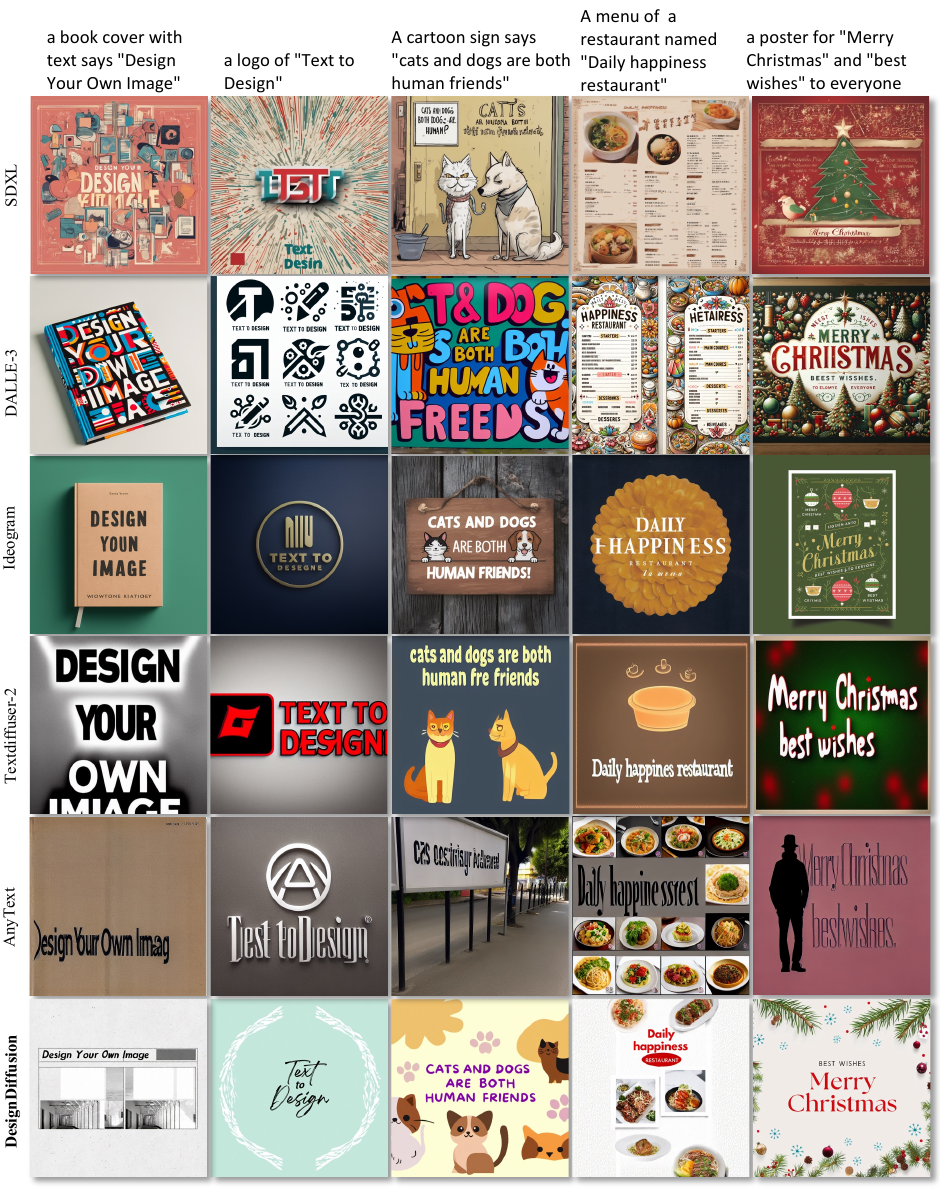}
  \vspace{-2em}
  \caption{Qualitative comparisons with previous state-of-the-art text-to-image generation and visual text rendering methods reveal that our DesignDiffusion produces more elegant and harmonious integrated visual and textual design images.}
  \label{fig:visual_compare}
\end{figure*}

\begin{table*}[t]
\caption{Ablation studies of the proposed framework. \textbf{Bold} and \underline{underline} denotes the first and second best performance. ``FT'' and ``PE'' represent whether to apply fine-tuning and prompt enhancement, respectively. }
\vspace{-1em}
\centering
\resizebox{1.0\textwidth}{!}{
\setlength\tabcolsep{10pt}
\begin{tabular}{cccc|ccccc}
\hline
FT & PE & $\mathcal{L}_{char}$ & SP-DPO & FID$\downarrow$ & Text Precision$\uparrow$ & Text Recall$\uparrow$ & Text F-score$\uparrow$ & Text Accuracy$\uparrow$\\ \hline
\gxmark & \gxmark & \gxmark & \gxmark& 45.10 & 0.517 & 0.461 & 0.488 & 0.241 \\
\cmark & \gxmark & \gxmark & \gxmark & 22.73 & 0.715 & 0.678 & 0.696 & 0.388\\
\cmark & \cmark & \gxmark & \gxmark & 19.89 & 0.862 & 0.799 & 0.829 & 0.572 \\
\cmark & \cmark & \cmark & \gxmark & \textbf{19.29}&\underline{0.881}  & \underline{0.831} & \underline{0.855} & \underline{0.616} \\
\cmark & \cmark & \cmark & \cmark & \underline{19.87} & \textbf{0.888} & \textbf{0.837} & \textbf{0.862} & \textbf{0.631} \\ 
\hline
\end{tabular}
}
\vspace{-1em}

\label{table:ablation_study}
\end{table*}

\subsection{Comparison with State-of-the-Art Methods}
\noindent\textbf{Quantitative Results.} We report the comparisons of the performance of design image generation in Table~\ref{table:quan_result}. We compare our method with SDXL~\cite{podell2023sdxl}, SD3~\cite{sd3}, FLUX.1~\cite{flux2023}, DeepFloyd-IF~\cite{deepfloyd}, GlyphControl~\cite{yang2024glyphcontrol}, Textdiffuser~\cite{chen2023textdiffuser}, Textdiffuser-2~\cite{textdiffuser-2}, and AnyText~\cite{tuo2023anytext}. The former four methods are open-source state-of-the-art text-to-image generation models. The latter three are state-of-the-art visual rendering or inpainting methods. Our DesignDiffusion outperforms other methods by a large margin with a FID score of 19.87, demonstrating the high quality of our generated design images. From the aspect of visual text quality, we report the OCR text precision, recall, F-score, and accuracy systematically and comprehensively. The comparison tells us that our DesignDiffusion gets a state-of-the-art visual text generation performance.
The performance of our method in both the image and text metrics indicates that DesignDiffusion establishes a new baseline in design image generation.

\noindent\textbf{Qualitative Results.} Visual comparison experiments are further conducted and the result is shown in Figure~\ref{fig:visual_compare}. We compare our method with state-of-the-art text-to-image models, \ie, SDXL~\cite{podell2023sdxl}, DeepFloyd-IF~\cite{deepfloyd}, DALLE-3~\cite{betker2023improving}, Ideogram~\cite{ideogram}, and state-of-the-art visual text rendering models, \ie, Textdiffuser~\cite{chen2023textdiffuser}, Textdiffuser-2~\cite{textdiffuser-2}, AnyText~\cite{tuo2023anytext}.
We observe that although state-of-the-art text-to-image models~(SDXL, Deepfloyd-IF, DALLE-3, Ideogram) can generate visually compelling images, they still struggle to generate accurate visual text. As for visual text rendering methods, we figure out that they generate visual text while brushing out the image content or reducing the aesthetics of the image seriously. We guess that since Textdiffuser~\cite{chen2023textdiffuser}, Textdiffuser-2~\cite{textdiffuser-2}, AnyText~\cite{tuo2023anytext} both rely on the text layout planning or strong text position fitting, their models overfit on visual text and overlook other visual contents. Besides, over-strong regularization of visual text area or other information limits the creativity of generating visual content. The visual comparisons demonstrate that design images generated by our DesignDiffusion are more accurate than those from state-of-the-art text-to-image models and more elegant than those generated from state-of-the-art visual text rendering models. 

\begin{figure*}[t]
  \centering
    \includegraphics[width=1.0\linewidth]{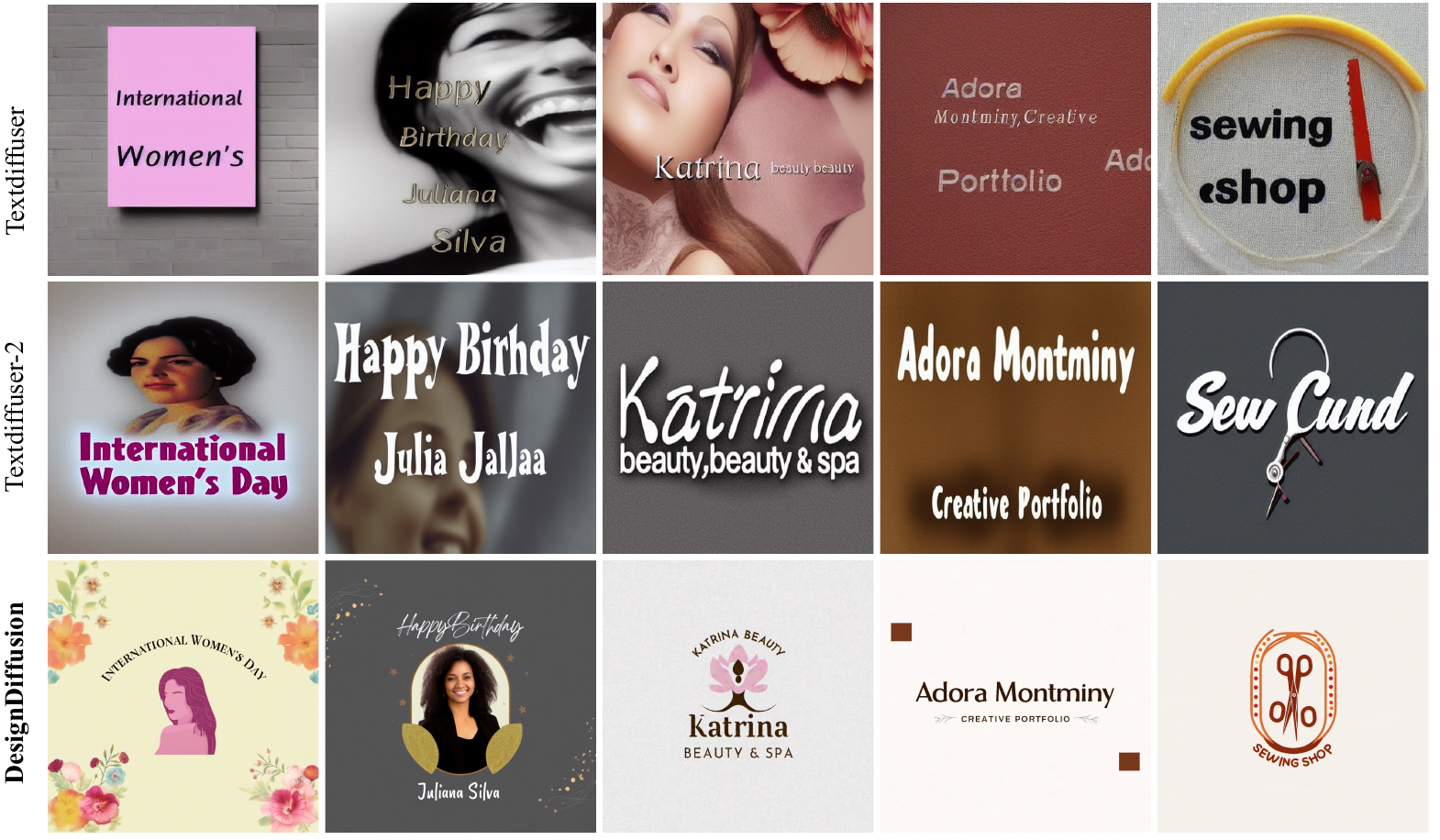}
    \vspace{-2em}
  \caption{Visual comparisons of the capabilities of automatic text layout planning from our model with those of planning by language models. Upon training, our DesignDiffusion has demonstrated the ability to generate flexible and well-organized text layouts effectively.}
  \vspace{-1em}
  \label{fig:visual_layout}
\end{figure*}

\noindent\textbf{User Studies.} We randomly generate 20 samples without any cherry pick compared with three methods, \ie, SDXL~\cite{podell2023sdxl}, Textdiffuser-2~\cite{textdiffuser-2}, and AnyText~\cite{tuo2023anytext}. The participants were asked four questions about various aspects of image and text evaluation, in terms of image aesthetic, visual text quality, layout aesthetics, and text-image matching. 
We collect about 20 results from participants and report the preference ratio of the four metrics in Table~\ref{table:user_study}. From the aspects of text quality, text layout aesthetic, and text-image matching, our DesignDiffusion significantly outperforms the competitors. It obtains comparable, slightly worse than SDXL~\cite{podell2023sdxl} in evaluating image aesthetic, \ie, 40\% v.s. 43\%. It may be because fine-tuning SDXL especially for the design image generation task loses a bit of aesthetic.

\noindent\textbf{Automatic Layout Planning.} DesignDiffusion is a one-stage method without any layout planning module. We visualize examples compared with layout-dependent methods~\cite{chen2023textdiffuser,textdiffuser-2} in Figure~\ref{fig:visual_layout}. 
Samples from others often have an oversized font or dull layout design, while our DesignDiffusion generates images with curved, colorful, and diverse text. The results show that our DesignDiffusion has a strong automatic, flexible, and reasonable layout planning capability.

\subsection{Ablation Studies}
To assess the effectiveness of the proposed components within our framework, we conduct ablation studies.
We explore various configurations, with the baseline being the original SDXL model. We denote fine-tuning on our dataset as ``FT'', the utilization of prompt enhancement, $\mathcal{L}_{char}$, and self-play DPO  as``PE'', ``$\mathcal{L}_{char}$'' and ``SP-DPO'', respectively. The ablation results are summarized in Table~\ref{table:ablation_study}, revealing the following notable insights.

\noindent\textbf{Fine-tuning enhances the quality of design image generation:} Fine-tuning SDXL with design images results in a significant improvement in FID. Despite this enhancement in image quality, achieving satisfactory accuracy in rendered text remains a challenge for existing diffusion models, even with extensive fine-tuning on large-scale datasets.

\noindent\textbf{Effect of prompt enhancement:} Our investigation highlights limitations of current Byte Pair Encoding (BPE) tokenization, particularly in providing precise character-level cues for the visual text rendering task. Incorporating prompt enhancement leads to a significant improvement across various metrics, notably enhancing visual text accuracy because of the introduced character-level information.

\noindent\textbf{Localized attention through $\mathcal{L}_{char}$:} By utilizing character segmentation masks to guide the cross-attention maps within the UNet architecture, our $\mathcal{L}_{char}$ loss facilitates focused attention on text regions. With $\mathcal{L}_{char}$, we observe a further enhancement in text precision, recall, F-score, and accuracy, emphasizing the effectiveness of localizing cross-attention maps using character segmentation masks.

\noindent\textbf{Enhanced visual text accuracy with SP-DPO:} Subsequent to fine-tuning, we integrate SP-DPO to further improve the generation quality. SP-DPO significantly enhances text quality across all text metrics. Regarding the slight decline in the FID score, this can be attributed to the specific implementation of SP-DPO, which relies heavily on the selection of winning and losing data. Currently, the winning-losing pairs are determined based on the text accuracy. If we incorporate image quality into the criteria for constructing the losing data, we anticipate that the FID score could improve further. Notably, our current approach does not visually compromise image quality while enhancing text quality. 

\section{Discussion and Conclusion}
\label{sec:conclusion}
In this paper, we introduce DesignDiffusion, a diffusion-based fine-tuning framework designed for synthesizing design images from textual descriptions. Unlike previous approaches that handle visual and textual elements independently, potentially limiting creativity and leading to style or color inconsistencies in style or color harmony, DesignDiffusion offers an end-to-end, one-stage diffusion-based solution. We explore including a distinctive character embedding and a character localization loss for enhancing visual text learning. Additionally, we employ SP-DPO fine-tuning strategy to improve the generation quality.  Our extensive experiments demonstrate that DesignDiffusion achieves state-of-the-art performance in design image generation, showcasing its effectiveness and robustness in addressing the complexities of visual text synthesis. We anticipate that our research will stimulate further exploration in design image generation, fostering innovation and advancement in this field.

{
    \small
    \bibliographystyle{ieeenat_fullname}
    \bibliography{main}
}

\clearpage

\appendix
\section*{Appendix}
\section{More Dataset Details}
Our dataset consists of images sourced from Google Image Search using design-related keywords such as logo, poster, flyer, cover, sign, brochure, banner, business card, website mockup, packaging, magazine layout, advertisement, infographic, product label, menu, invitation, certificate, and presentation slide. We use the LLaVA 1.6-34B vision-language model to extract visual text and generate captions.

We apply several filters to ensure the quality of our dataset:
\begin{itemize}
    \item \textbf{Text Length}: Images with visual text longer than 150 characters are excluded.
    \item \textbf{Resolution}: Images with resolutions lower than 768×768 pixels are removed.

    \item \textbf{Aspect Ratio}: Images with aspect ratios outside the range of 0.25 to 4.0 are excluded.
    
    \item \textbf{Aesthetic Quality}: Images with an aesthetic score lower than 4.5 are filtered out.
\end{itemize}
After filtering, our dataset includes about 1 million high-quality training images and 5,000 testing images.

Moreover, we elaborate further on the criteria that define ``high quality'' in our context:
\begin{itemize}
    \item \textbf{Design-Specific Content}: The images in our dataset are specifically collected for their relevance to design purposes. This includes images featuring visual texts commonly found in design contexts. By focusing on these types of images, we ensure that the dataset is highly relevant to the task of text-to-design image generation.

    \item \textbf{Enhanced Image Captions}: The captions for the images are generated using the state-of-the-art, open-source large vision-language model, LLaVA 1.6-34B. This model provides detailed and contextually accurate descriptions of the images, which enhances the quality of the dataset by offering rich textual information that aligns closely with the visual content.
\end{itemize}
These criteria collectively define the ``high quality'' nature of our dataset, ensuring that it is both relevant and robust for the task at hand.

Our dataset is specifically curated for design image generation, setting it apart from Mario-10M~\cite{chen2023textdiffuser}, LAION-Glyph~\cite{yang2024glyphcontrol}, and AnyWord-3M~\cite{tuo2023anytext}, which primarily focus on visual text rendering. The key differences include:
\begin{itemize}
    \item \textbf{Content Composition}: Our dataset comprises images that seamlessly integrate both visual and textual elements to reflect design aesthetics. In contrast, the other datasets often focus solely on text, such as tables or isolated sentences, without the broader visual context or the integration of design elements.

    \item \textbf{Design Focus}: Our dataset is specifically tailored for design contexts, encompassing a wide variety of design assets, including logo, poster, flyer, cover, sign, brochure, banner, business card, website mockup, packaging, magazine layout, advertisement, infographic, product label, menu, invitation, certificate, and presentation slide. The other datasets, however, generally focus on the presence of text within images, without specifically addressing the diverse and specialized design contexts that our dataset targets.
    Overall, we provide detailed statistics of the composition of our design dataset compared with the existing visual text rendering dataset in Table~\ref{table:dataset_stat1}.
\end{itemize}

\begin{table}
    \small
    \centering
    \caption{\small Statistics of the composition of our design dataset.}
    \vspace{-1em}
   \resizebox{1.0\linewidth}{!}{
    \begin{tabular}{l|c|c|c|c|c}
    \hline
      Dataset & img count & img type& mean lines/img  & chars/words count & unique chars/words \\\hline
    AnyWord-3M & 3M& img w/ text & 4.13 & 6.35M & 695.2K \\
    Ours & 1M& design img& 4.71 & 16.72M & 251.1K \\\hline
    \end{tabular}}
    \vspace{-.5em}
    \label{table:dataset_stat1}
\end{table}

\section{OCR Tool Comparison}
\label{appendix:ocr_tool}
\begin{figure*}[h]
  \centering
  \includegraphics[width=1.0\linewidth]{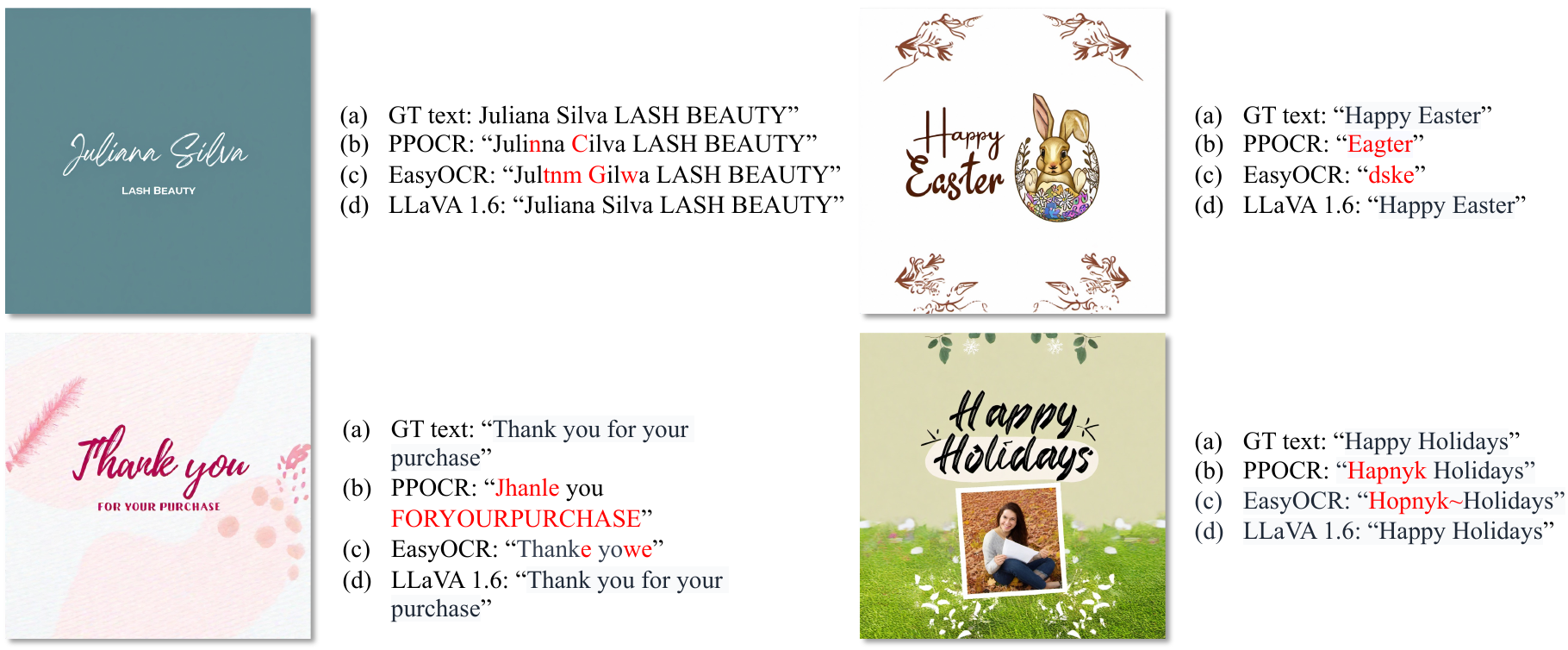}
  \vspace{-1.5em}
  \caption{Examples for comparing the OCR capability for detecting text in design images. \textcolor{red}{Red} part denotes wrong detection. LLaVA 1.6 gets the best OCR accuracy for extracting text in design images.}
  \label{fig:ocr_compare}
\end{figure*}

To evaluate the visual text quality in generated design images, we need an OCR tool to extract the visual text rendered in the image. We demonstrate that two popular OCR tools~( PPOCR~\cite{du2020pp}, EasyOCR~\cite{easyocr}) often fail to extract visual text in design images. After deep investigations, we find that current large-scale multi-modality models do a better job than those OCR tools. Here, we give examples to show that LLaVA 1.6 gets a higher OCR accuracy of extracting text in design images than PPOCR~\cite{du2020pp}, EasyOCR~\cite{easyocr}), as shown in Fig.~\ref{fig:ocr_compare}.

\section{Image Captioning by LLaVA}
\label{appendix:caption}
\begin{figure*}[h]
  \centering
  \includegraphics[width=1.0\linewidth]{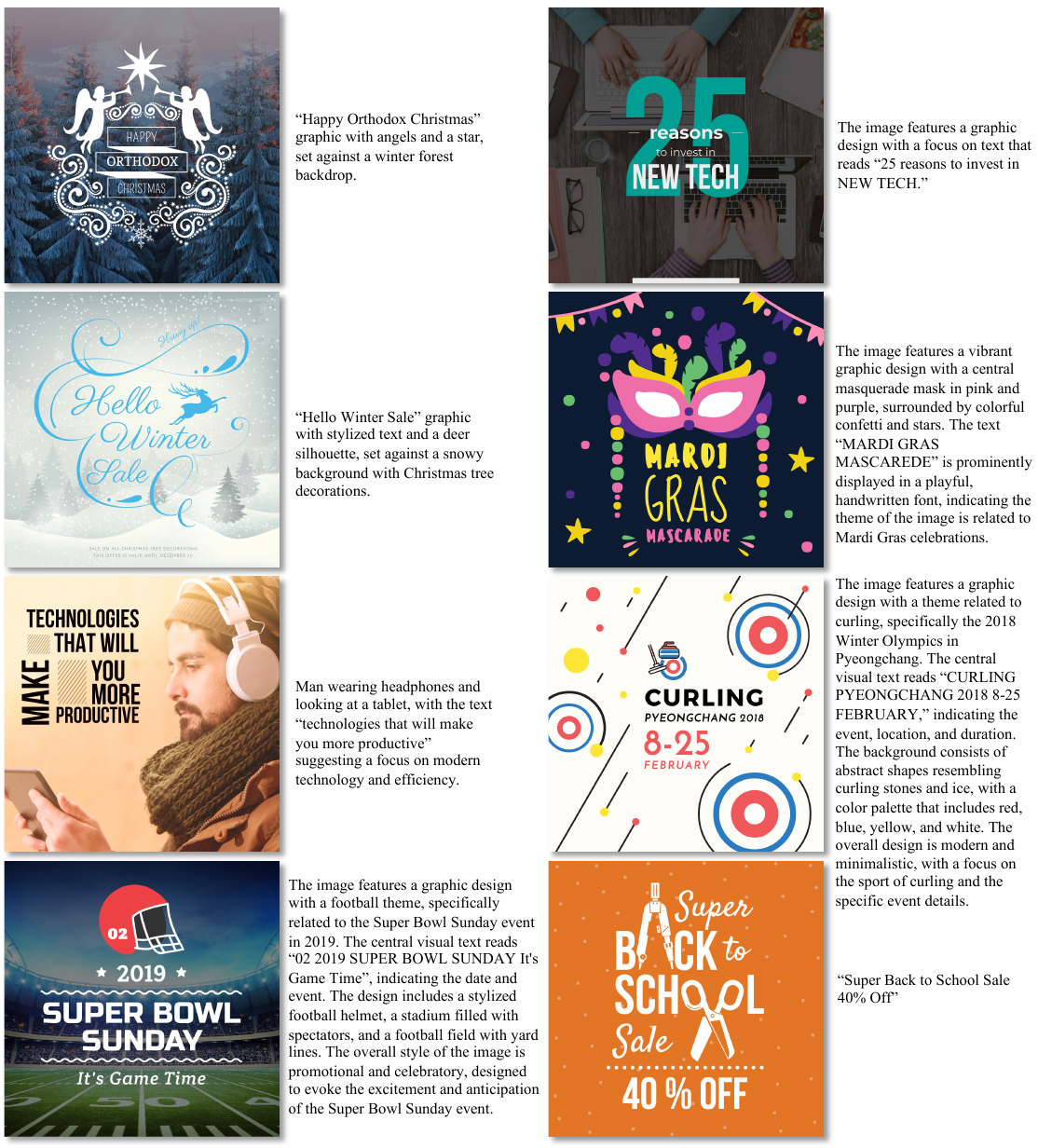}
  \vspace{-1.5em}
  \caption{Examples of image-prompt pair from the design image dataset.}
  \label{fig:dataset_example}
\end{figure*}

In our design image dataset, we adopt LLaVA-1.6-34B~\cite{liu2024llavanext} to describe the input image. Considering that we have obtained the text annotations of these images, we design a comprehensive prompt to encourage LLaVA-1.6 to perform image captioning:
\begin{tcolorbox}[colframe=blue!10, colback=yellow!10]
    Please provide a concise caption for the image, detailing the text shown in the image and key image elements from both holistic and detailed perspectives. If there is text content, ensure that any extracted text content in your output is enclosed in double quotes. Do not include irrelevant information or subjective comments in your output. When given the image's text content, you should output a definite, unequivocal, correct, and objective caption that incorporates all the text information. \textcolor{gray}{<optional if text available>} The text content in this image is <text place-holder>.
\end{tcolorbox}
After image captioning by LLaVA-1.6-34B, we further ensure that the returned caption includes the visual text information. If not containing complete text information, we append the lost text information into the returned caption by applying a random manual-designed template. For example, `with a text of ``<text place-holder>''', or `with words saying ``<text place-holder>''', etc. We visualize some examples of our design image dataset in Fig.~\ref{fig:dataset_example}.

\begin{figure}[t]
  \centering
  \includegraphics[width=\linewidth]{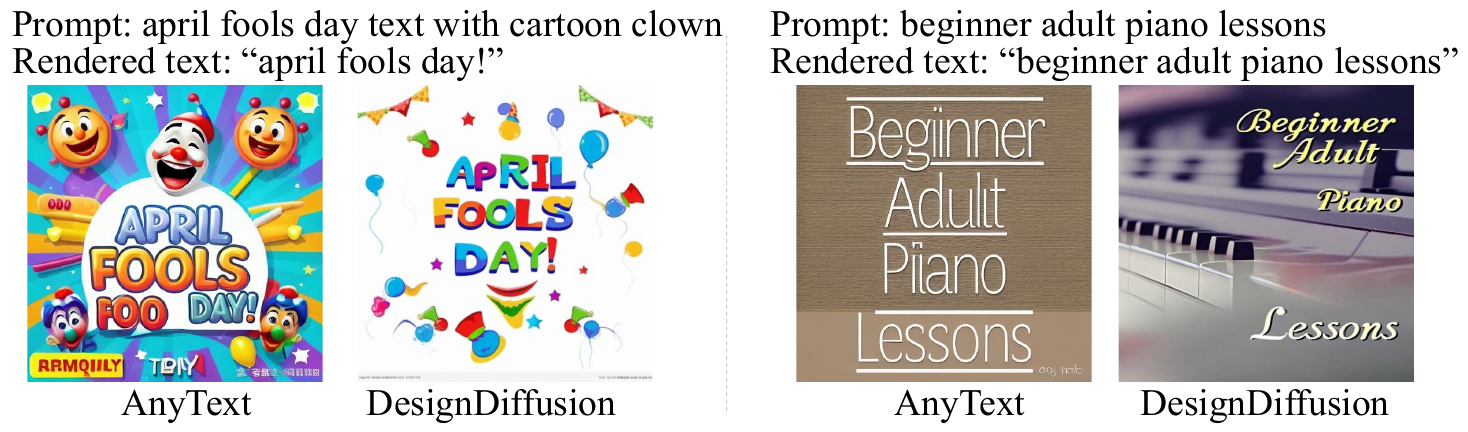}
  \vspace{-2.2em}
  \caption{Qualitative comparisons on AnyWord-3M.}
  \label{fig:anytext}
\end{figure}

\begin{figure*}[h]
  \centering
  \includegraphics[width=1.0\linewidth]{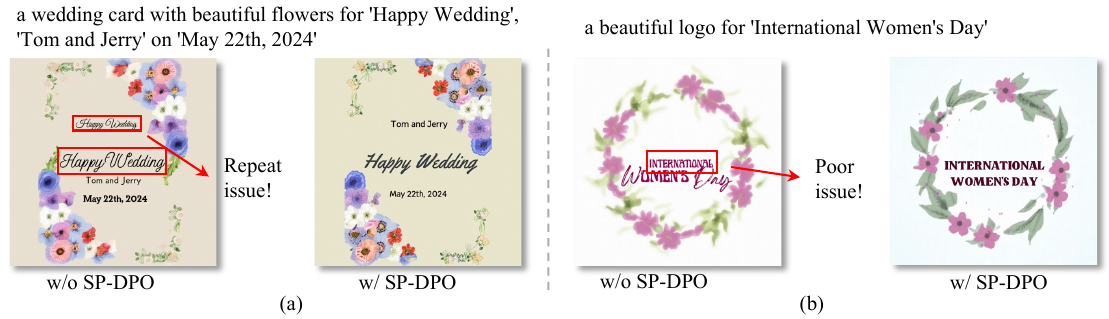}
  \vspace{-1.5em}
  \caption{Effect of SP-DPO. With SP-DPO, our model reduces the phenomenon of repeated text and poor text quality while keeping the overall image quality.}
  \label{fig:effect_dpo}
\end{figure*}

\section{Effect of SP-DPO}
We have shown the quantitative result of SP-DPO in the ablation studies in the main paper. Our ablation studies indicate that SP-DPO significantly enhances text quality across all text metrics. Regarding the slight decline in the FID score, this can be attributed to the specific implementation of SP-DPO, which relies heavily on the selection of winning and losing data. Currently, the winning-losing pairs are determined based on the text accuracy of synthetic images. If we incorporate image quality into the criteria for constructing the losing data, we anticipate that the FID score will improve further.

Notably, our current approach does not visually compromise image quality while enhancing text quality. In Figure~\ref{fig:effect_dpo}, we provide samples to illustrate the positive impact of SP-DPO on text quality without detracting from the overall image quality.

\section{Diverse Generation}

\begin{figure*}[h]
  \centering
  \includegraphics[width=1.0\linewidth]{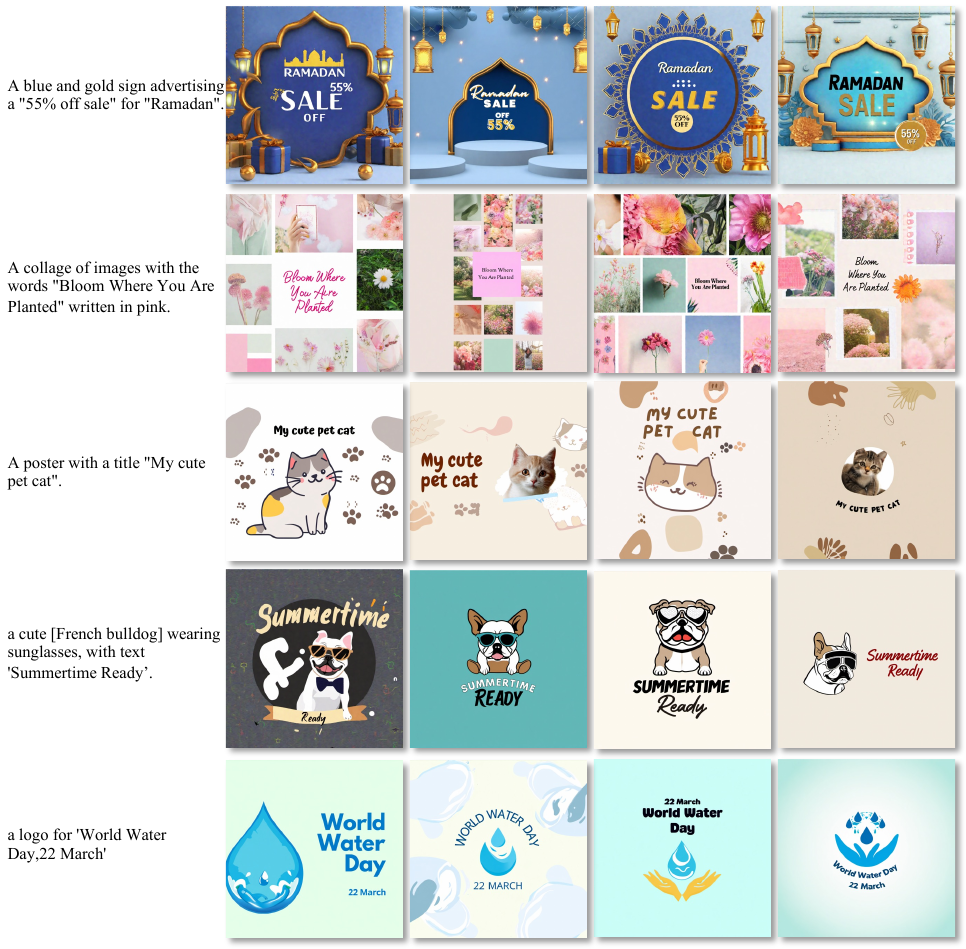}
  \vspace{-1.5em}
  \caption{Diverse samples generated by DesignDiffusion. Our DesignDiffusion generates high-quality diverse design images with a one-stage framework.}
  \label{fig:diverse_sample}
\end{figure*}

Diverse samples generated by DesignDiffusion are shown in Fig.~\ref{fig:diverse_sample}. Given one prompt, we generate four images for a more clear demonstration. We observe that DiffusionDiffusion generates design images with different image styles, text positions, and image-text corporations.

\begin{figure*}[h]
  \centering
  \includegraphics[width=1.0\linewidth]{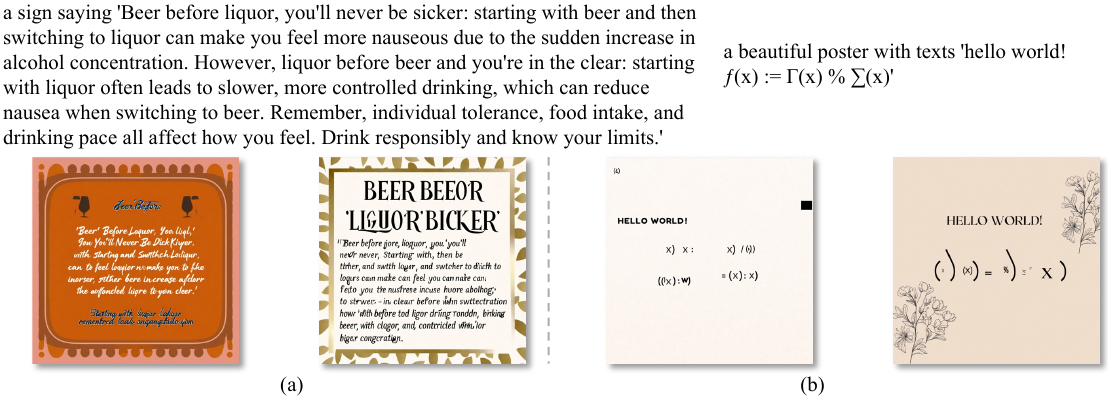}
  \vspace{-1.5em}
  \caption{Failure cases. (a) Handling Very Long Visual Text. (b) Rare Formula Symbols.}
  \label{fig:failure_case}
\end{figure*}

\section{Comparisons on AnyWord-3M} Although our work focuses on design image generation, we further conduct experiments on the public AnyWord-3M dataset~\cite{tuo2023anytext}, training our DesignDiffusion for 40k iterations with a batch size of 128. 
This corresponds to only 1/6 of the training epochs used for AnyText. Table~\ref{table:quanti_res} and Figure~\ref{fig:anytext} present quantitative and qualitative comparisons, respectively. Our results demonstrate that DesignDiffusion outperforms AnyText in terms of both visual text rendering accuracy and image quality.

\begin{table}[t]
    \small
    \centering
    \caption{\small Quantitative comparison on AnyWord-3M. }
    \vspace{-1em}
    \resizebox{1.0\linewidth}{!}{
    \setlength\tabcolsep{15pt}
    \begin{tabular}{l|c|c|c|c}
    \hline
    Method & Training Epochs & Sen. ACC↑ & NED↑ & FID↓  \\ \hline
    AnyText~\cite{tuo2023anytext} & 10 & 0.6588 & 0.8568 & 35.87 \\
    DesignDiffusion & 1.7 & \textbf{0.6922} & \textbf{0.8719} & \textbf{33.98 }\\
    \hline
    \end{tabular}
    }
    \label{table:quanti_res}
\end{table}

\section{Failure Cases}
In our experiments, we discovered some failure cases in which current models may also struggle to generate these images. We show two cases in Figure~\ref{fig:failure_case}. (a) Handling Very Long Visual Text: Our model may struggle to generate accurate visual text when the text length is exceptionally long.
(b) Rare Formula Symbols: The model has difficulty generating rare formula symbols, primarily because these symbols are infrequently or never encountered in the training data. We believe that identifying and analyzing these failure cases is crucial for guiding future improvements in the task of design image generation.

\end{document}